\documentclass[letterpaper, 10 pt, conference]{ieeeconf}  

\IEEEoverridecommandlockouts                              

\usepackage{times}
	\usepackage{graphicx,times}
	\usepackage{subfigure}
\usepackage{amsmath}
\usepackage{amssymb}
\usepackage{mathrsfs}
\usepackage{multirow}
\usepackage{booktabs}
\usepackage{makecell}
\usepackage{array}
\usepackage{url}
\usepackage{cite} 
\usepackage{mathptmx} 
\newcolumntype{P}[1]{>{\centering\arraybackslash}p{#1}}
\newcolumntype{M}[1]{>{\centering\arraybackslash}m{#1}}

\title{\LARGE \bf
NormNet: Scale Normalization for 6D Pose Estimation in Stacked Scenarios
}

\author{En-Te Lin, Wei-Jie Lv, Ding-Tao Huang and Long Zeng*
\thanks{* corresponding author.}
\thanks{ En Te Lin, Wei Jie Lv, Ding Tao Huang and Long Zeng are with the Department of Advanced Manufacturing, Shenzhen International Graduate School, Tsinghua University, Shenzhen, China (e-mail: linet22@mails.tsinghua.edu.cn; lwj19@mails.tsinghua.edu.cn; hdt22@mails.tsinghua.edu.cn; zenglong@sz.tsinghua.edu.cn).}%
}

\begin{document}

\maketitle
\thispagestyle{empty}
\pagestyle{empty}

\begin{abstract}

Existing Object Pose Estimation (OPE) methods for stacked scenarios are not robust to changes in object scale. This paper proposes a new 6DoF OPE network (NormNet) for different scale objects in stacked scenarios. Specifically, each object's scale is first learned with point-wise regression. Then, all objects in the stacked scenario are normalized into the same scale through semantic segmentation and affine transformation. Finally, they are fed into a shared pose estimator to recover their 6D poses. In addition, we introduce a new Sim-to-Real transfer pipeline, combining style transfer and domain randomization. This improves the NormNet's performance on real data even if we only train it on synthetic data. Extensive experiments demonstrate that the proposed method achieves state-of-the-art performance on public benchmarks and the MultiScale dataset we constructed. The real-world experiments show that our method can robustly estimate the 6D pose of objects at different scales.
\end{abstract}

\section{INTRODUCTION}

Accurate 6D object pose estimation in stacked scenarios is a crucial prerequisite for vision-guided robot grasping in industry and logistics applications. A stacked scene is a pile of different objects randomly clustered together, e.g., in bin-picking \cite{bregier2017symmetry} and table organization \cite{liu2021ocrtoc}. It is a challenging task due to the objects' variety, senor noise, and heavy occlusion. 

Existing methods for 6D OPE can be roughly divided into traditional and learning-based methods. Traditional methods \cite{lowe1999object}, \cite{rothganger20063d} extract and match the hand-designed feature from the scene and object's model to recover the 6D pose. The performance of these methods drops significantly in stacked scenes with similar objects. With the development of deep learning and simulation techniques \cite{bregier2017symmetry}, \cite{kleebergerlarge, landau2015simulating, dai2022domain}, the learning-based method has dominated the 6D OPE task in stacked scenarios.
Some works \cite{dong2019ppr}, \cite{kleeberger2020single} directly regress the 6DoF pose with networks, and others \cite{he2020pvn3d}, \cite{zeng2021parametricnet} recover 6D pose by least-squares fitting between predicted objects keypoints and pre-defined keypoints in the CAD models.
However, these methods estimate all objects in the same OPE pipeline, neglecting the impact of object scale, which may result in limited performance on some objects with special scales.

\begin{figure}[t]
	\centering
		\includegraphics[width=1.0\columnwidth]{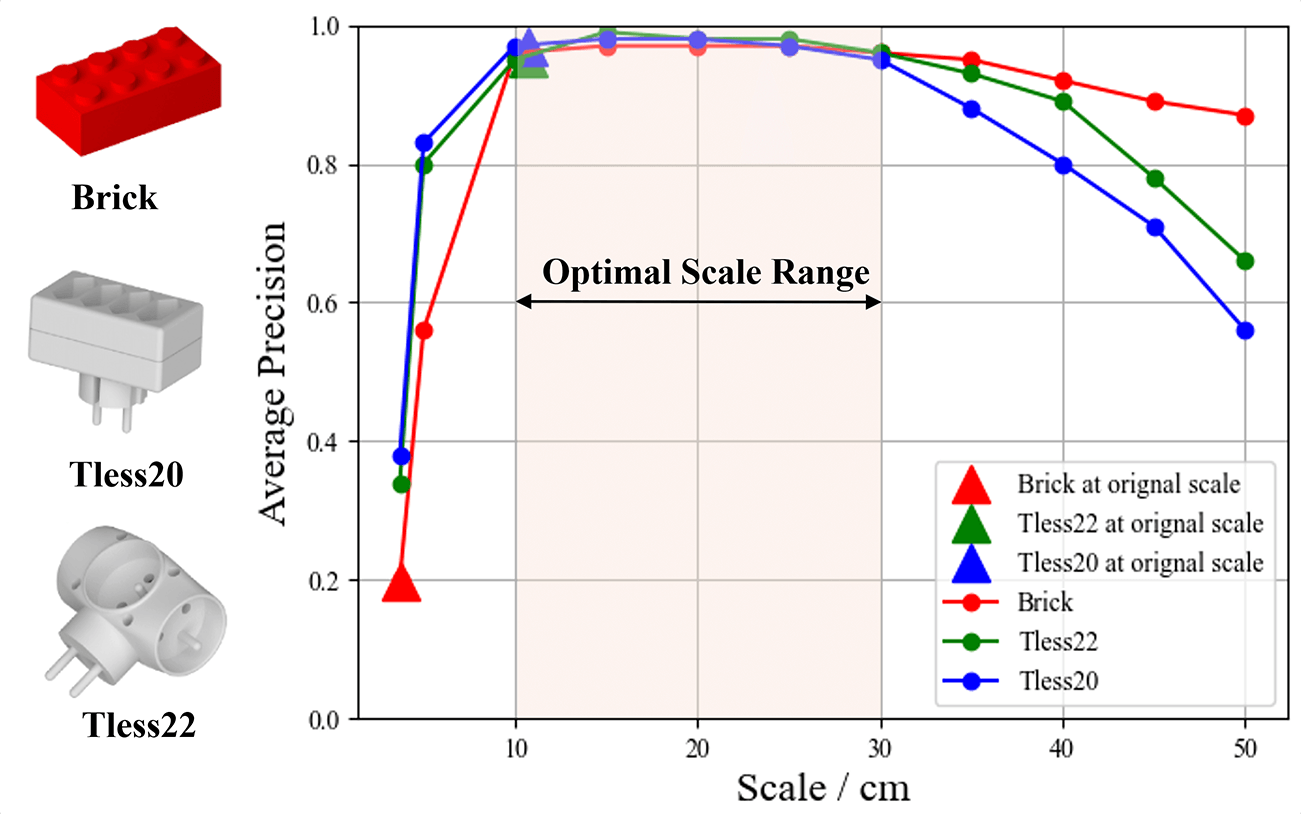}
	\caption{The performance of PPR-Net is affected by object scale. “Optimal Scale Range” refers to the scale range of the AP curve above 0.95 and the performance of the brick object drops significantly at its original scale.}
	\label{fig:AP_curve}
        \vspace{-10pt}
\end{figure}

To analyze the specific impact of the object scale on the OPE task, we conduct an experiment on objects of different scales using a representative OPE method for stacked scenes, PPR-Net \cite{dong2019ppr}, and the result is shown in Fig. \ref{fig:AP_curve}. The scale is defined as the diameter of the object’s smallest bounding sphere. We selected three objects from the Fraunhofer IPA Bin-Picking dataset \cite{kleebergerlarge} and rescaled their model to obtain objects of different scales (5-50 cm). Then we respectively generated the simulated stacked scenario datasets for objects of each scale to train and evaluate the network. It is obvious that the network can achieve the best performance within the optimal scale range (10-30 cm) and the object of too small or too large scale will incur a significant decline in performance. We think the reason is that the fixed feature receptive field and hyperparameters (e.g., the bandwidth \cite{zeng2021ppr} of the mean-shift clustering algorithm for instance segmentation ) in the same OPE pipeline might be only optimal for the objects within a certain scale range.

\begin{figure}[t]
	\centering
		\includegraphics[width=0.6\columnwidth]{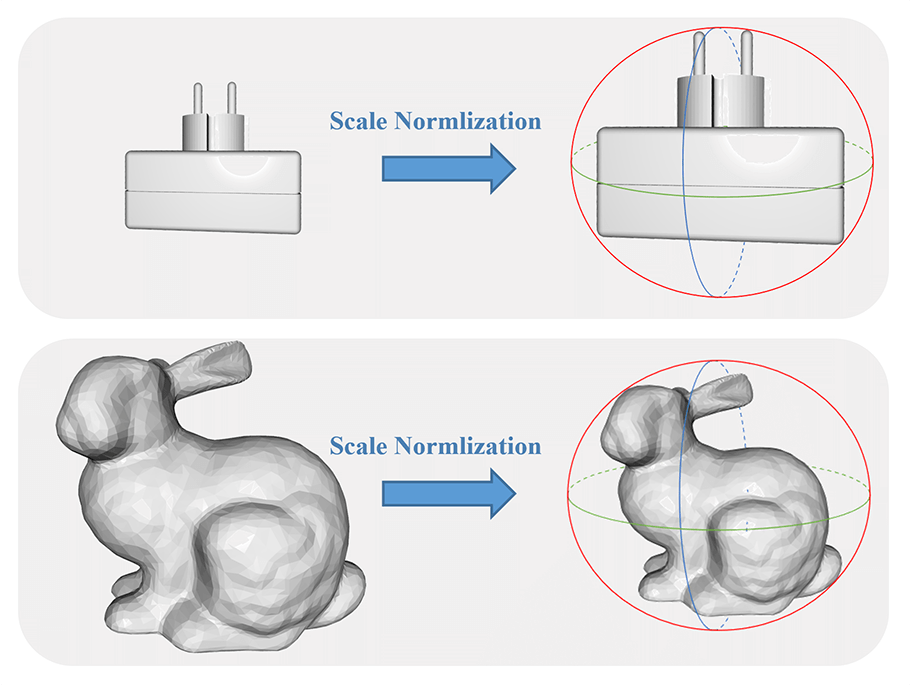}
	\caption{Objects of different scales are normalized into the SNCS.}
	\label{fig:scale_normalization}
        \vspace{-10pt}
\end{figure}

To eliminate the impact of object scale, we propose a new 6DoF pose estimation network for objects of different scales in stacked scenarios, denoted as NormNet. Inspired by the works \cite{singh2018analysis}, \cite{wang2019normalized}, we define a Scale Normalized Coordinate Space (SNCS) in which all objects share the same scale and the scale in the SNCS is optimal for the feature receptive field and the hyperparameters in the OPE pipeline. The core of NormNet is a scale normalization module, where the objects of different scales are normalized into the SNCS, as shown in Fig. \ref{fig:scale_normalization}. Subsequently, the objects in the SNCS are fed into a shared pose estimator for pose estimation.

Furthermore, NormNet is trained on synthetic data so the Sim-to-Real gap will diminish the model’s performance on real data. The Sim-to-Real gap contains noise patterns and depth missing, e.g., the reflective or thin surfaces in some small objects will result in depth missing in the real depth image \cite{cao2023two,zhang2018deep}. Previous works \cite{zeng2021ppr,kleeberger2020single} are only for noise-related problems and can't handle the depth missing. To further bridge the Sim-to-Real gap, we propose a new learning-based Sim-to-Real transfer pipeline. Specifically, the depth missing can be considered as a style of the real data, so we employ a style transfer model to generate the depth missing on the synthetic data; Then, we apply domain randomization to the synthetic data to cover the real noise pattern.

We evaluate the NormNet on three datasets, i.e., Sil\'eane Dataset \cite{bregier2017symmetry}, Parametric dataset \cite{zeng2021parametricnet}, and MultiScale dataset.  On the Sil\'eane Dataset, compared with the state-of-the-art methods OP-NET and PPR-Net, our method improves mAP by 10\% and 7\%, respectively. On the Parametric dataset, our method improves mAP by 24\% and 16\% compared with PPR-Net and ParametricNet, respectively. We construct the MultiScale dataset which consists of seven objects with significant scale variations. On the MultiScale dataset, our method outperforms PPR-Net by 21\% in the mAP.

In summary, the main contributions of our work are:

\begin{itemize}
	\item A new 6DoF OPE network with scale normalization module is proposed for objects of different scales in stacked scenarios.
	\item A new learning-based Sim-to-Real transfer pipeline is proposed to further bridge the Sim-to-Real gap.
	\item A new MultiScale dataset is constructed and evaluated.
\end{itemize}

\section{Related Work}

\subsection{Pose estimation method}

6D OPE methods can be roughly classified into traditional and learning-based methods. Traditional methods tackle the 6D OPE by template or feature matching. In the template-based method \cite{hinterstoisser2013model,hinterstoisser2011multimodal}, the objective is to find the most similar labeled templates for the objects. The templates are constructed by projecting the 3D model from different viewpoints. A similarity score is used to match the objects with templates. In feature-based methods \cite{lowe1999object,rothganger20063d}, the local features are extracted and matched between objects and 3D models to recover pose. However, these methods are not robust to stacked scenes. Recently, some learning-based methods \cite{xiang2017posecnn,wang2019densefusion} first extract high-dimensional features, and directly regress the 6DoF pose of objects by the network. To address the non-linear problem in the rotation space, some works \cite{peng2019pvnet,he2020pvn3d} propose to predict corresponding keypoints and recover the 6D pose by least-squares fitting. Others \cite{su2015render,sundermeyer2018implicit,tulsiani2015viewpoints} simplify rotation prediction into a classification task. To tackle the category explosion in parametric shapes, Zeng et al. \cite{zeng2021parametricnet} treat a shape template as a category and propose a point-wise regression and sparse keypoint recovery for 6D OPE. Although learning-based methods have achieved impressive results, they all have neglected the impact of object scale and estimated the objects of different scales in the same pipeline, which will lead to limited performance for some objects with special scales.

\subsection{Sim-to-Real transfer method}

Sim-to-Real transfer aims to bridge the Sim-to-Real gap between synthetic and real data. The Sim-to-Real transfer methods can be roughly divided into domain adaption and domain randomization. Most domain adaptation methods aim to reduce the feature representation differences between source and target domains. SqueezeSegV2 \cite{wu2019squeezesegv2}, ePointDA \cite{zhao2021epointda} compress the point clouds into 2D and perform domain transfer. Some works \cite{saleh2019domain,bousmalis2017unsupervised} achieve domain adaption by adversarial training, and others \cite{achituve2021self} by a shared feature encoder. 
In contrast, Domain randomization alleviates the Sim-to-Real gap by introducing randomness to synthetic data generation. Some works \cite{prakash2019structured,tremblay2018training,yue2019domain} generate synthetic data with sufficient variation by randomizing the simulation setting, e.g., background, illumination. Others \cite{zeng2021ppr,kleeberger2020single} domain randomize the generated data by adding noise and blurring. However, the domain randomization is only for noise-related problems and is unable to handle stylistic differences. In this paper, we combine the domain randomization with a style transfer model to further bridge the Sim-to-Real gap.

\section{Method}

\begin{figure}[t]
	\centering
		\includegraphics[width=0.9\columnwidth]{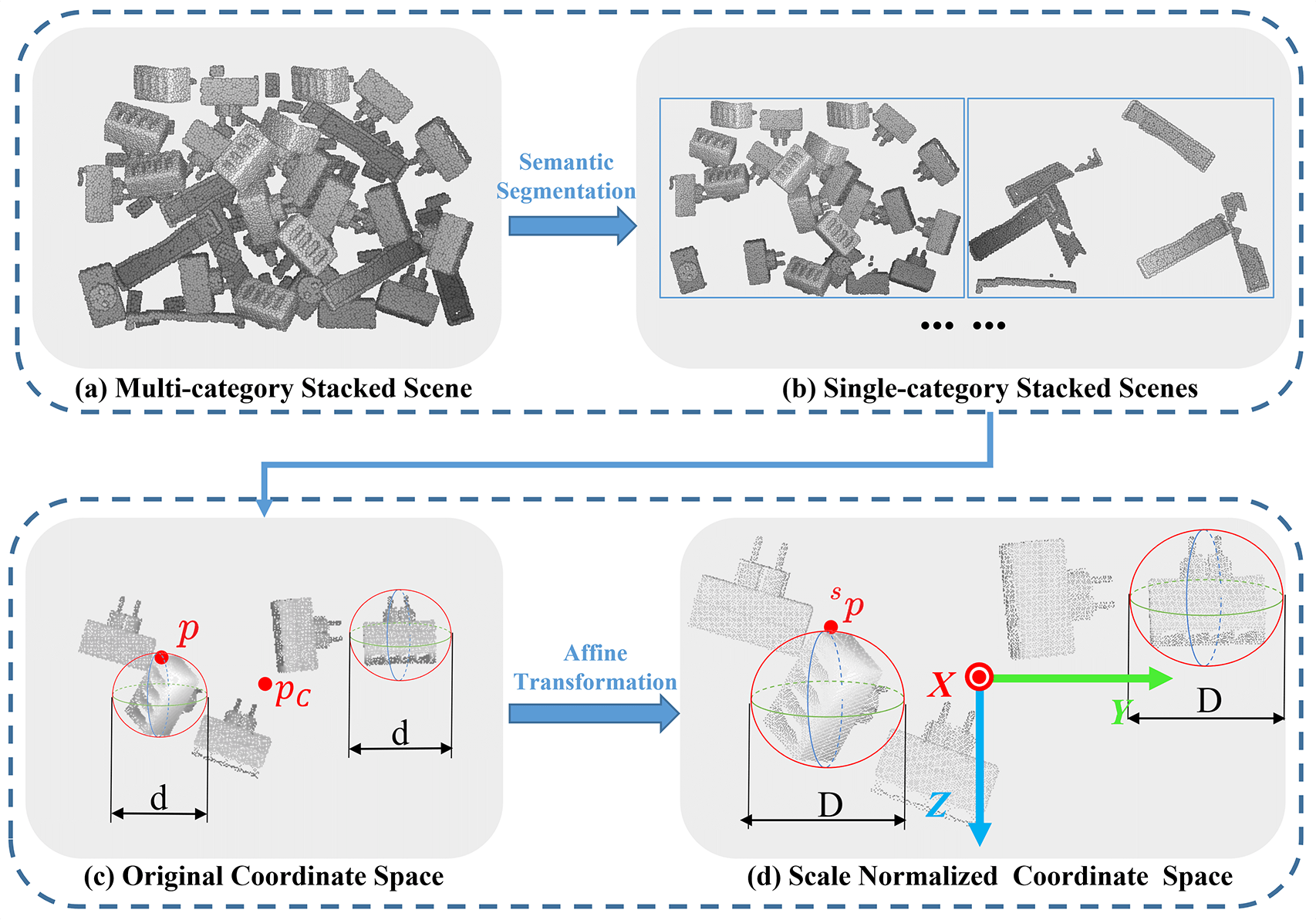}
	\caption{Scale Normalization on multi-category stacked scene. The multi-category stacked scene is semantically segmented into several single-category stacked scenes and then each scene is transformed into the SNCS, respectively. }
        \vspace{-10pt}
	\label{fig:SNCS}
\end{figure}

\begin{figure*}[t]
	\centering
		\includegraphics[width=2.0\columnwidth]{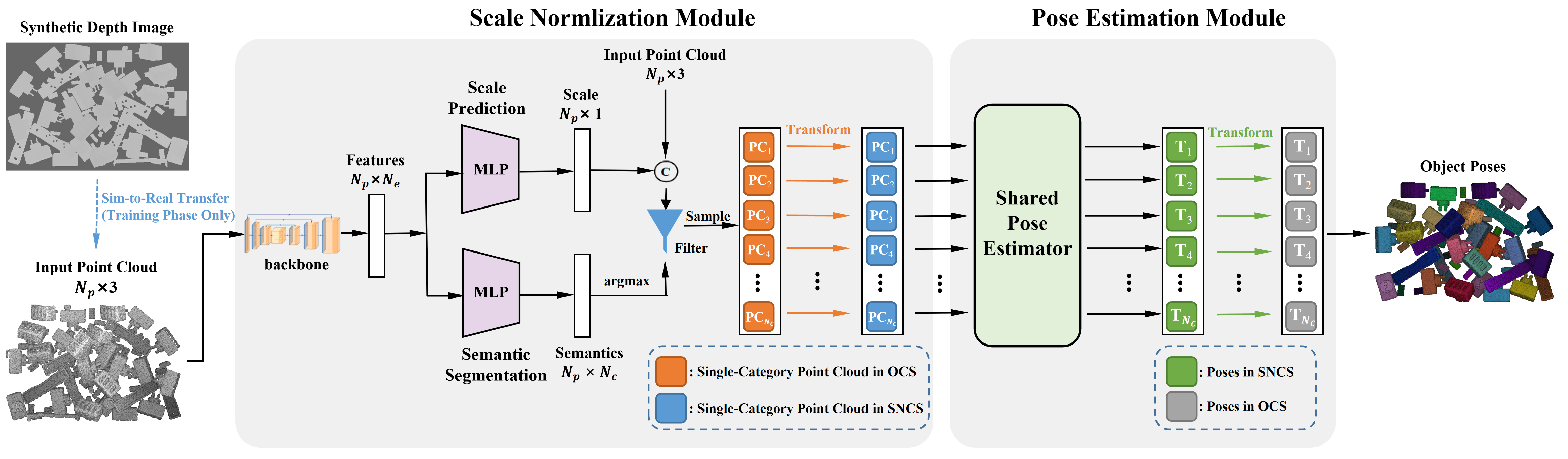}
	\caption{NormNet architecture, where OCS and SNCS represent Original Coordinate Space and Scale Normalized Coordinate Space, $PC_i$ represents the point cloud of $i$th single-category stacked scene, $T_i$ represents the poses estimation result of $PC_i$.}
	\label{fig:overview_of_NormNet}
         \vspace{-10pt}
\end{figure*}

\subsection{Scale Normalized Coordinate Space (SNCS)} 

The stacked scenes can be divided into multi-category and single-category stacked scenes according to the number of object categories in the scene, and a multi-category stacked scene can be composed of several single-category stacked scenes. For each single-category stacked scene, we define the SNCS as a 3D space in which all objects share the same scale $D$ within the optimal scale range. The scale normalization is the process of transforming the stacked scene from its Original Coordinate Space (OCS) to the SNCS. Specifically, given a point cloud of a multi-category stacked scene in the OCS (Fig. \ref{fig:SNCS}(a)), it can first be semantically segmented into several single-category stacked scenes (Fig. \ref{fig:SNCS}(b)). As shown in Fig. \ref{fig:SNCS}(c), we denote 
$p_c$ as the center of a single-category stacked scene. Since all objects in the scene share the same scale $d$, the single-category stacked scene can be transformed into the SNCS by affine transformation which uniformly scales each object’s scale to $D$ and centers the scene in the SNCS:
\begin{align}
	^{\text{\tiny{\textit{S}}}}p = \frac{D}{d}(p-p_c),
\label{tansform2SNCS}
\end{align}
where $p$ is the point on the object in the OCS,$^{\text{\tiny{\textit{S}}}}p$ is its corresponding point in the SNCS, as shown in Fig. \ref{fig:SNCS}(d).

\subsection{Architecture Design} 

\textbf{Method overview}. The architecture of NormNet is shown in Fig. \ref{fig:overview_of_NormNet}. The network consumes an unordered point cloud of a stacked scene of $N_p\times 3$. The point cloud is semantically segmented and transformed into the SNCS by the scale normalization module. Then each single-category point cloud in the SNCS is respectively fed into a shared pose estimator for instance segmentation and pose estimation. Finally, the object poses are obtained by transforming the poses in the SNCS back to the OCS. The network jointly learns the prediction of scale, semantics, pose and visibility in the stacked scene:
\begin{align}
	L = \lambda_{1} \cdot L_{scale} + \lambda_{2} \cdot L_{semantic} + \lambda_{3} \cdot L_{pose} + \lambda_{4} \cdot L_{visibility},
\end{align}
where $\lambda_{1}$, $\lambda_{2}$, $\lambda_{3}$, and $\lambda_{4}$ are loss weights to ensure that the four losses in $L$ are roughly equally weighted.

\textbf{Scale Normalization Module}. As shown in Fig. \ref{fig:overview_of_NormNet}, the extracted features of size $N_p \times N_f$ are fed into two separated MLPs. 
The first MLP predicts the scale of object to which each point belongs. 
The second MLP produces the semantic prediction $\hat{C}$ of size $N_p \times N_c$, $N_c$ is the number of object categories in the scene, and element $\hat{c}_{ij}$ is the probability for $i$th point belonging to $j$th class. 
The input point cloud is concatenated with the point-wise scale prediction and then segmented semantically into several point clouds of single-category stacked scenes  $\{PC_i\}_{i=1}^{N_C}$ by the semantic prediction. 
To avoid imbalance of the point number across the category, we uniformly sample each point cloud to the same number $N_s$. 
For each point cloud in $\{PC_i\}_{i=1}^{N_C}$, the scale $d$ is obtained  by averaging the point-wise predicted scales.
Then $\{PC_i\}_{i=1}^{N_C}$ are respectively transformed into SNCS according to Eq. \ref{tansform2SNCS}. 
We supervise the scale and the semantic prediction with L1 and cross-entropy loss:
\begin{align}
	L_{scale} &= \frac{1}{N_p}\sum _{i=1}^{N_p}{\left\|s_i-\hat{s_i}\right\|}, \\
	L_{semantic} &= \frac{1}{N_p}\sum _{i=1}^{N_p}{c_{i}log(\hat{c}_{i})},
\end{align}
where $s_i$ and ${\hat{s}_i}$ are point-wise predicted scale and scale label, $c_{i}$ is the one-hot representation of ground true class label.

\begin{figure}[t]
	\centering
		\includegraphics[width=1.0\columnwidth]{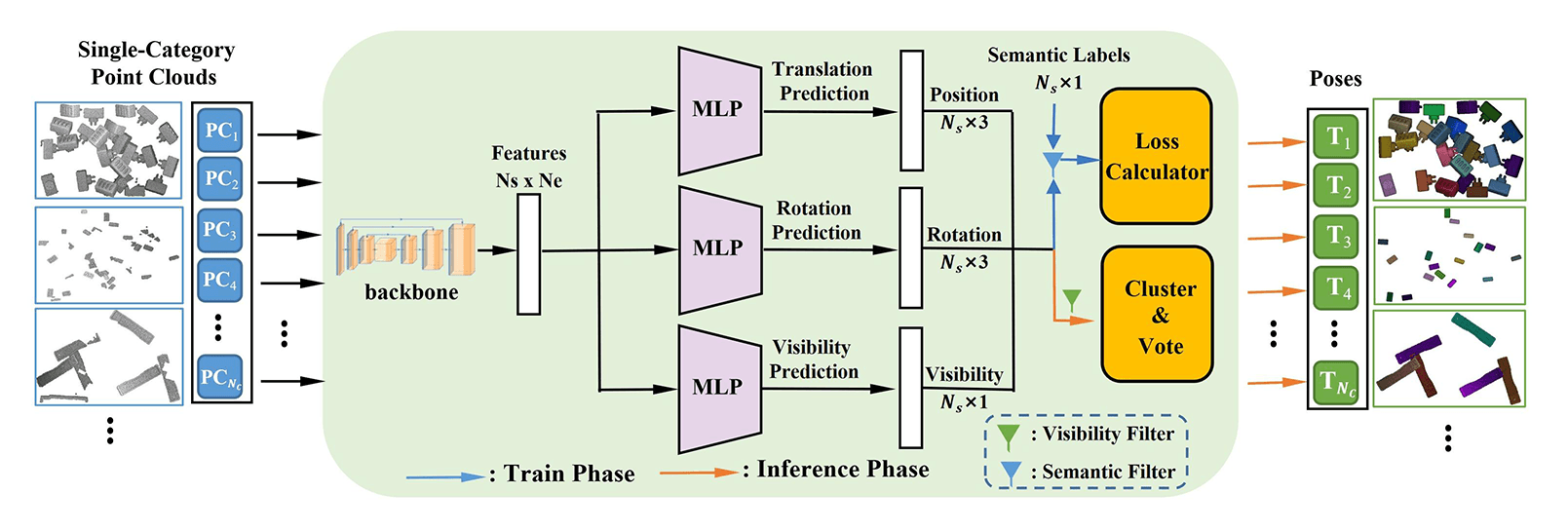}
	\caption{Shared Pose Estimator, based on point-wise regression and Hough voting scheme.}
	\label{fig:pose_estimator}
    \vspace{-10pt}
\end{figure}

\textbf{Pose Estimation Module}. As shown in Fig. \ref{fig:pose_estimator}, this module takes the point cloud of single-category stacked scene $PC_j$ as input. Similar to PPR-Net\cite{dong2019ppr}, the extracted features of size $N_s \times N_f$ are fed into three separated MLPs to produce translation, rotation and visibility prediction, respectively. Pose can be represented as a rigid transformation $T$ by combining the prediction of translation and rotation. In the training phase, we modify the loss function by adding a semantic filter to avoid interference of points in other categories. The total losses are averaged by all categories:
\begin{align}
	L_{pose} = \frac{1}{N_c}\frac{1}{N_s}\sum _{j=1}^{N_c}\sum _{i=1}^{N_s}dist({T}^j_i, \hat{T}^j_i)\mathbb{I}(c^j_i=j), \\
	L_{visibility}= \frac{1}{N_c}\frac{1}{N_s}\sum _{j=1}^{N_c}\sum _{i=1}^{N_s}{\left\|V^j_i-\hat{V}^j_i\right\|}\mathbb{I}(c^j_i=j),
\end{align}
where $dist(\cdot)$ is a pose distance proposed by Romain Br\'egier et al. \cite{bregier2018defining}, 
$\hat{T}^j_i$ and ${T}^j_i$ are point-wise predicted pose and pose label, $\hat{V}^j_i$ and $V^j_i$ are point-wise predicted visibility and visibility label, $\mathbb{I}$ is an indicating function equates to 1 only when the semantic label of $i$th point in $PC_j$ belong to $j$th class, and 0 otherwise.  

In the inference phase, the point-wise prediction of translation and rotation are filtered by a visibility threshold $T_v$. Then the pose of each instance in the SNCS can be obtained by gathering the point-wise pose prediction results with clustering algorithms \cite{comaniciu2002mean} and voting in pose space. Finally, we obtain each instance’s pose in the OCS by fixing the rotation and applying an affine transformation to the translation, similar to Eq. \ref{tansform2SNCS}:
\begin{align}
	\text{\large{\textit{t}}} = \frac{D}{d}{^{\text{\tiny{\textit{S}}}}\text{\large{\textit{t}}}} + p_c ,
\end{align}
where $\text{\large{\textit{t}}}$ and $^{\text{\tiny{\textit{S}}}}\text{\large{\textit{t}}}$ are translation vector in the OCS and the SNCS.

\begin{figure}[t]
	\centering
		\includegraphics[width=1.0\columnwidth]{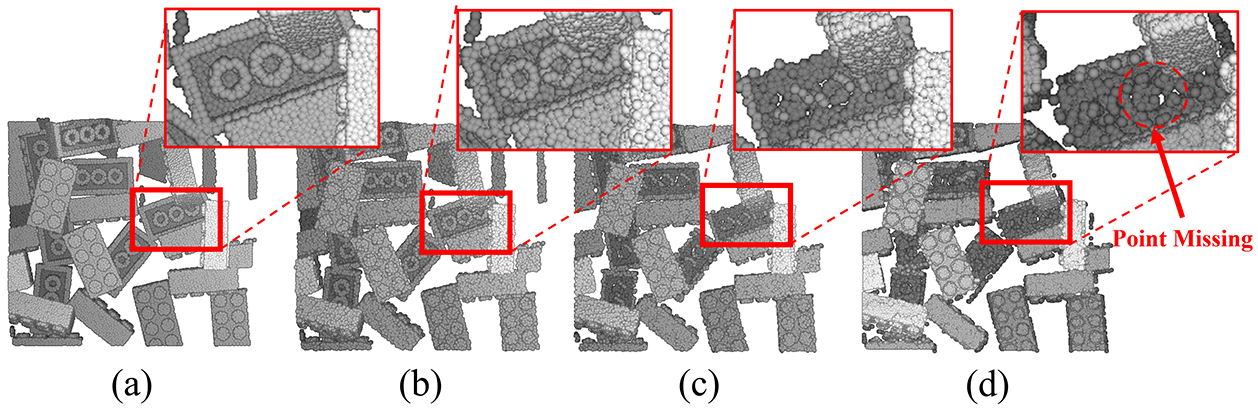}
	\caption{Sim-to-Real gap on point cloud. (a) Synthetic point cloud. (b) Synthetic point cloud with domain randomization. (c) Transferred point cloud. (d) Real point cloud.}
	\label{fig:fig_shape_miss}
 \vspace{-10pt}
\end{figure}

\subsection{Sim-to-Real transfer} \label{section:SRT}
The Sim-to-Real gap between the synthetic point cloud and the real point cloud is shown in Fig. \ref{fig:fig_shape_miss}. Compared with the real point cloud (Fig. \ref{fig:fig_shape_miss}(d)), the synthetic point cloud (Fig. \ref{fig:fig_shape_miss}(a)) does not contain any noise and point missing. As shown in Fig. \ref{fig:fig_shape_miss}(b), domain randomization can introduce noise into the synthetic point cloud but can not generate point missing.

\begin{figure}[t]
	\centering
		\includegraphics[width=1.0\columnwidth]{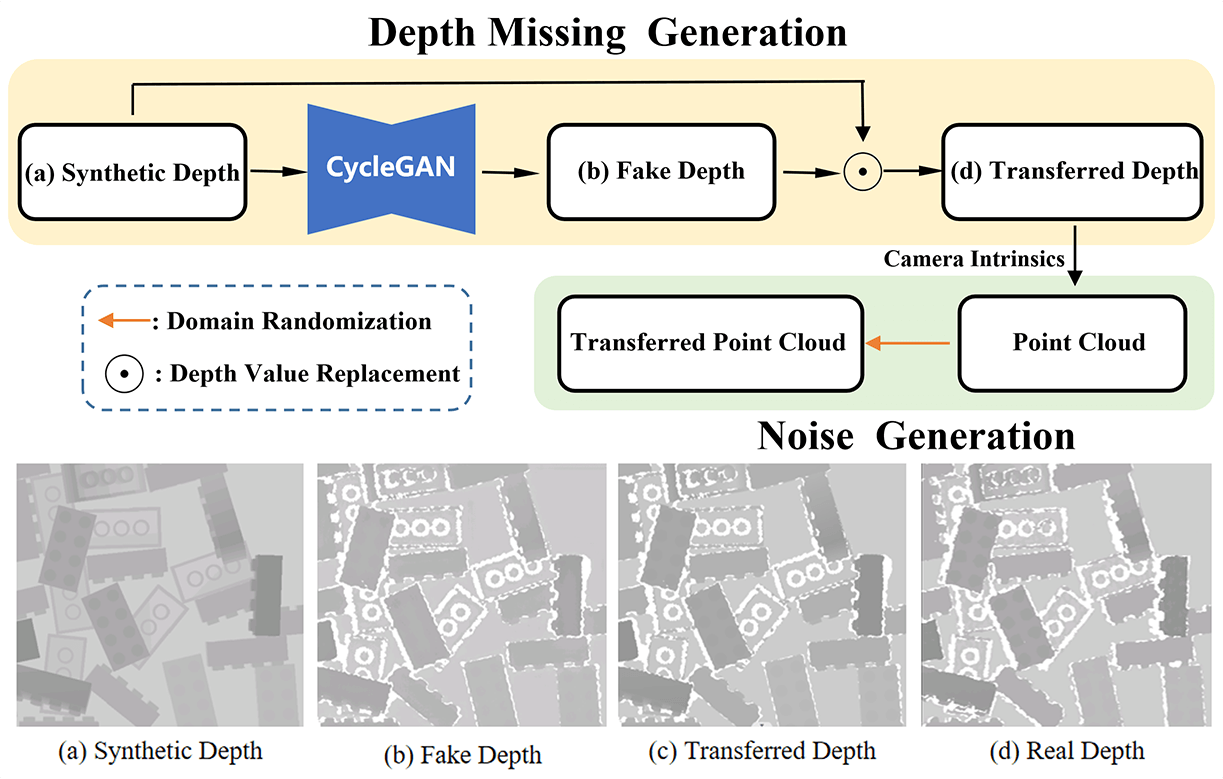}
	\caption{The pipeline of learning-based Sim-to-Real method. For comparison, the synthetic depth is reconstructed from the real depth.}
	\label{fig:fig_DT_pipeline}
 \vspace{-10pt}
\end{figure}

To bridge the Sim-to-Real gap, we proposed a new learning-based Sim-to-Real transfer pipeline as shown in Fig. \ref{fig:fig_DT_pipeline}. The point missing on the point cloud is the depth missing in the depth image (Fig. \ref{fig:fig_DT_pipeline}(d)), since the point cloud is converted from the depth image. The depth missing can be considered as a style of the real depth image, so we employ a style transfer model, CycleGAN, to generate the depth missing on the synthetic depth image. Specifically, we trained the CycleGAN with unpaired synthetic and real depth images to learn the distribution of the depth missing. A synthetic depth image (Fig. \ref{fig:fig_DT_pipeline}(a)) rendered by Blender \cite{kleebergerlarge} is then fed into the trained CycleGAN to generate a fake depth image with depth missing (Fig \ref{fig:fig_DT_pipeline}(b)). Though similar to real depth image, the depth values in the fake depth are inaccurate. Therefore, we replace the depth values in the fake depth image with those from the synthetic depth image to obtain a transferred depth image (Fig. \ref{fig:fig_DT_pipeline}(c)). With camera intrinsic, we convert the transferred depth image to the point cloud. Finally, we apply the domain randomization to the point cloud by adding the standard normal distribution noise and obtain the transferred point cloud for training, as shown in Fig. \ref{fig:fig_shape_miss}(c). The transferred point cloud is similar to the real point cloud, which improves the generalization ability of the network in the real data.

\subsection{Training and implementation}
In the implementation, 16,384 points are sampled for each scene using furthest point sampling and each point contains only camera coordinate information.
Pointnet ++ \cite{qi2017pointnet++} is chosen as the feature extraction backbone and $N_f$ = 128, $T_v$ = 0.50, $\lambda_{1}$ = 2, $\lambda_{2}$ = 20, $\lambda_{3}$ = 0.2, and $\lambda_{4}$ = 50 are set by default.
The scale $D$ in SNCS is chosen as 20 cm according to experiment on different objects.
Our network is prototyped with Pytorch 1.8 and optimized by the Adam optimizer with an initial learning rate of 0.001.
On the Sil\'eane Dataset and the Parametric dataset, we train NormNet for 100 epochs with batchszie of 8 on an RTX4090 GPU. On the MultiScale dataset, we train NormNet for 100 epochs with batchsize of 2 on two RTX4090 GPUs.

\section{Experiment}

\begin{figure}[b]	
	\centering
	\subfigure[TN06]{	
		\begin{minipage}[b]{0.20\columnwidth}
			\includegraphics[width=0.98\columnwidth, height=1.5cm]{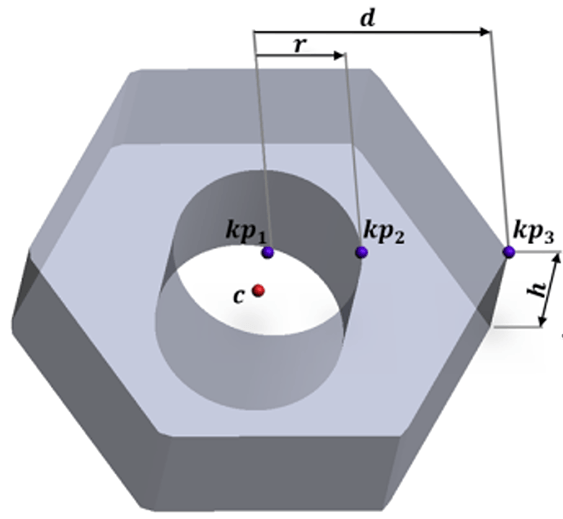}
		\end{minipage}
		\label{fig:fig7a}
		}
	\subfigure[TN42]{\centering
		\begin{minipage}[b]{0.23\columnwidth}
			\includegraphics[width=0.98\columnwidth, height=1.5cm]{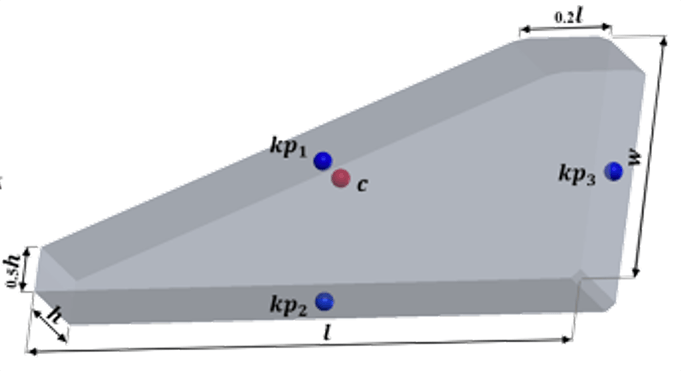}
		\end{minipage}
		\label{fig:fig7b}
		}		
	\subfigure[Rightangle]{\centering
		\begin{minipage}[b]{0.20\columnwidth}
			\includegraphics[width=0.98\columnwidth, height=1.5cm]{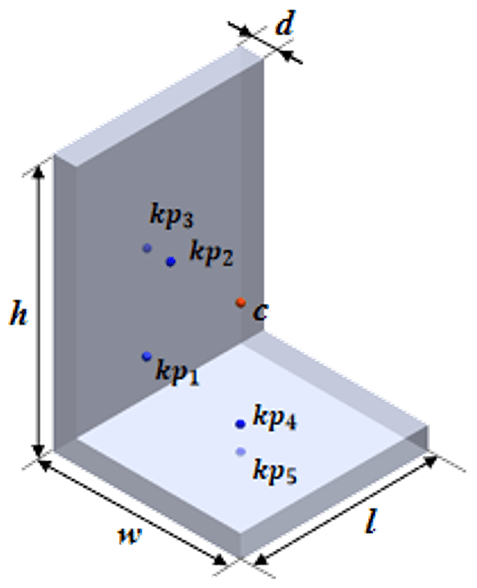}
		\end{minipage}
		\label{fig:fig7c}
		}					
		\subfigure[Semicircle]{\centering
		\begin{minipage}[b]{0.23\columnwidth}
			\includegraphics[width=0.98\columnwidth, height=1.5cm]{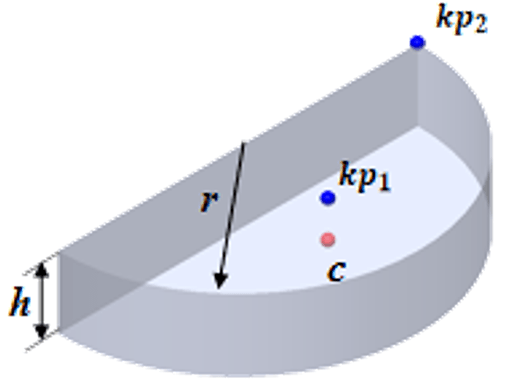}
		\end{minipage}
		\label{fig:fig7d}
		}				
	\caption{Parametric dataset shape parameters and keypoints.}
	\label{fig:Parametirc_shape}
  \vspace{-10pt}
\end{figure}

\begin{figure*}[t]
	\centering
		\includegraphics[width=2.0\columnwidth]{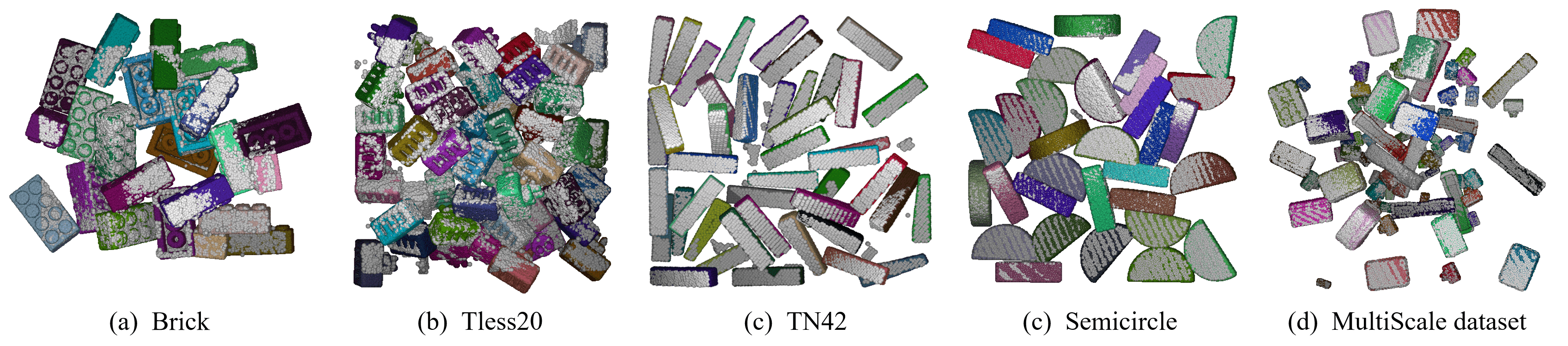}
	\caption{Performance on Sil\'eane dataset, Parametric dataset and MultiScale dataset. The estimated poses are overlapped with the point cloud.}
	\label{fig:fig_performance}
  \vspace{-10pt}
\end{figure*}

\subsection{Datasets and evaluation metrics} \label{section:dataset_metrics}

\textbf{Sil\'eane dataset}. The stacked scenes in the Sil\'eane dataset are single-category and close to the actual situation. We Select six objects in the Sil\'eane dataset, i.e., Brick, Candlestick (C.stick), Pepper, Bunny, Tless20 and Tless29, to evaluate the performance of NormNet.

\textbf{Parametric dataset}. The stacked scenes in the Parametric dataset are single-category and synthetic. The Parametric dataset has two test set: L-dataset and G-dataset to evaluate the learning and generalization ability, respectively. We augment the Parametric dataset with two new and common shape templates in the industry, i.e., Rightangle and Semicircle, and generate the L- and G- datasets the same way as ParametricDataset. As shown in Fig. \ref{fig:Parametirc_shape}, we select four shape templates in the Parametric dataset, i,e, Rightangle, Semicircle, TN06, TN42, to evaluate the performance of NormNet.

\begin{figure}[b]
	\centering
		\includegraphics[width=1.0\columnwidth]{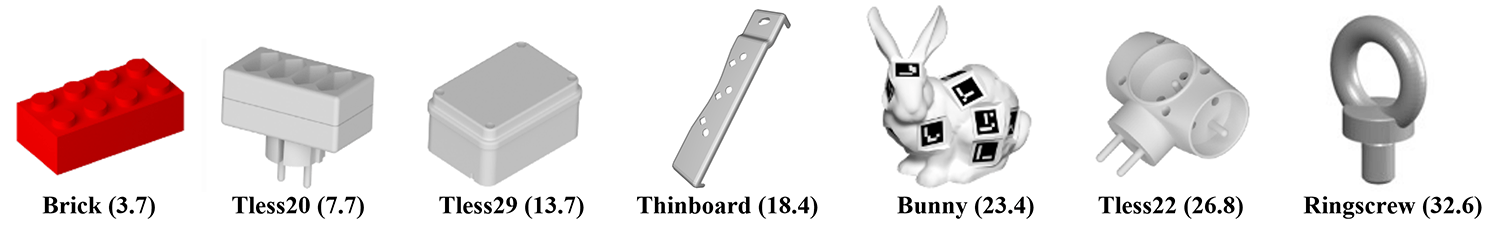}
	\caption{Objects of MultiScale dataset. The scales are shown within the brackets, units in cm.}
	\label{fig:MultiScale_datset}
  \vspace{-10pt}
\end{figure}

\textbf{MultiScale dataset}. The stacked scenes in the MultiScale dataset are multi-category and synthetic. As shown in Fig. \ref{fig:MultiScale_datset}, The MultiScale dataset contains seven types of objects, with scales ranging from 3.7-32.6 cm, and most of them are selected from the Fraunhofer IPA Bin-Picking dataset. For each stacked scene, we randomly select 3-4 types of objects and drop them into a bin to generate a stacked scene similar to the Sil\'eane dataset. We evaluate the performance of NormNet on all objects.

\textbf{Metircs}. The pose estimation metric proposed by Bregier \cite{bregier2017symmetry} is adopted. Individual instances with visibilities greater than 50\% are relevant for retrieval. The overall performance of pose estimation is measured by AP, which is the area under the precision-recall curve. For a parametric shape, the evaluation metric adopts mAP, which is the average of the APs of all part objects. 

\subsection{Comparison Result}
Some qualitative examples of NormNet are shown in Fig. \ref{fig:fig_performance}. The quantitative comparison and evaluation are given below.

\begin{table}[t]
\caption{Evaluation on Sil\'eane dataset, where OP-Net$_1$ is OP-Net with $Lori_1$ and $PP$ and OP-Net$_2$ is OP-Net with $Lori_2$ and $PP$ \cite{kleeberger2020single}.}
\resizebox{\linewidth}{!}{
\begin{tabular}{@{}cccccccc@{}}
\toprule
Object               & Brick         & C.stick      & Pepper        & Bunny         & Tless20       & Tless22         \\ \midrule
PPF PP \cite{drost2010model,bregier2017symmetry}     & 0.13          & 0.22         & 0.12          & 0.37          & 0.23          & 0.12          \\
LINEMOD PP \cite{hinterstoisser2013model,bregier2017symmetry} & 0.39          & 0.49         & 0.03          & 0.45          & 0.31          & 0.21          \\
PPR-Net \cite{dong2019ppr,zeng2021ppr}     & 0.47          & 0.98         & 0.98          & \textbf{0.99} & \textbf{0.93}          & 0.92          \\
OP-Net$_1$ \cite{kleeberger2020single}     & 0.42          & 0.97         & 0.98           & 0.94          & 0.88          & 0.86             \\
OP-Net$_2$ \cite{kleeberger2020single}        & 0.80          & 0.96         & 0.93          & 0.76          & 0.58          & 0.55          \\
\textbf{NormNet}     & \textbf{0.83} & \textbf{1.0} & \textbf{0.99} & \textbf{0.99} & \textbf{0.93} & \textbf{0.93} \\ \bottomrule
\end{tabular}
\label{tabel:Sileane}
}
\vspace{-10pt}
\end{table}

\textbf{Evaluation on Sil\'eane dateset}. NormNet is trained on the synthetic data from the Fraunhofer IPA Bin-Picking dataset, which has been transferred as the way mentioned in Sec. \ref{section:SRT}, and is evaluated on the Sil\'eane dataset. As shown in table \ref{tabel:Sileane}, NormNet outperforms the state-of-the-art methods on the Sil\'eane dataset. Specifically, on the average of APs on the six objects, while NormNet has no post-processing steps in the evaluation process, e.g., duplicate removal or ICP, it outperforms traditional non-learning methods, i.e., PPF PP, LINEMOD+ PP, by over 63\%; it also outperforms learning-based methods with post-processing steps, i.e., OP-Net$_1$ and OP-Net$_2$, by over 10\%. Compared with PPR-Net, the performance of NormNet is better, where APs of Brick, Candlestick, Tless20 and Tless22 are increased by 36\%, 2\%, 1\% and 1\%, respectively, while APs of Bunny and Tless20 are the same. Based on the above analysis, NormNet has stronger learning ability than other state-of-the-art methods.

\begin{figure*}[t]
	\centering
		\includegraphics[width=2.0\columnwidth]{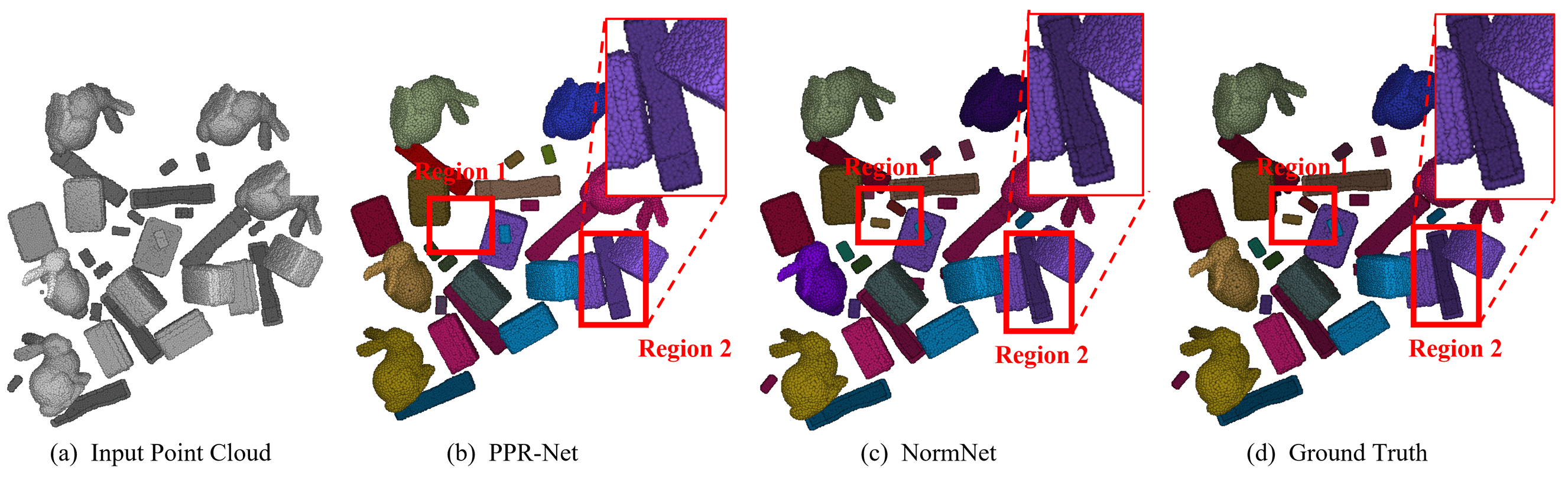}
	\caption{Qualitative results on the MultiScale dataset. In (b), PPR-Net misses the brick object in Region 1 and produces an incorrect pose prediction (upside down) for the thinboard in Region 2. In (c), NormNet with the scale normalization module performs better.}
	\label{fig:fig_compare}
\end{figure*}

\begin{table}[t]
\caption{Evaluation on Parametric dataset, where L is evaluation on L-dataset and G is evaluation on G-dataset}
\resizebox{\linewidth}{!}{
\begin{tabular}{@{}ccccccccc@{}}
\toprule
\multirow{2}{*}{Method} & \multicolumn{2}{c}{TN06}      & \multicolumn{2}{c}{TN42}      & \multicolumn{2}{c}{Rightangle} & \multicolumn{2}{c}{Semicircle} \\ \cmidrule(l){2-9} 
                        & L             & G             & L             & G             & L              & G              & L              & G             \\ \midrule
PPR-Net                 & 0.80          & 0.79          & 0.39          & 0.28          & 0.84           & 0.66           & 0.87           & 0.88          \\
ParametricNet \cite{zeng2021parametricnet}           & 0.94          & 0.93          & 0.52          & 0.51          & 0.88           & 0.74           & 0.82           & 0.83          \\
\textbf{NormNet}        & \textbf{0.97} & \textbf{0.96} & \textbf{0.98} & \textbf{0.98} & \textbf{0.91}  & \textbf{0.83}  & \textbf{0.89}  & \textbf{0.91} \\ \bottomrule
\end{tabular}
\label{tabel:Parametric}
}
\vspace{-7pt}
\end{table}

\begin{table}[t]
\caption{Evaluation on MultiScale dataset}
\resizebox{\linewidth}{!}{
\begin{tabular}{@{}cccccccc@{}}
\toprule
Method           & Brick         & Tless20       & Tless29       & Thinboard     & Bunny         & Tless22       & Ringscrew     \\ \midrule
PPR-Net          & 0.47          & 0.75          & 0.85          & 0.42          & 0.93          & 0.91          & 0.91          \\
\textbf{NormNet} & \textbf{0.95} & \textbf{0.98} & \textbf{0.97} & \textbf{0.97} & \textbf{0.99} & \textbf{0.98} & \textbf{0.97} \\ \bottomrule
\end{tabular}
\label{tabel:multiscale}
}
\vspace{-9pt}
\end{table}

\textbf{Evaluation on Parametric dataset}. All methods first learn all objects in the training set and are tested by the L-dataset and G-dataset. As shown in table \ref{tabel:Parametric}, it is obvious that NormNet outperforms PPR-Net and ParametricNet on all shape templates. Specifically, in the average of mAPs, NormNet outperforms PPR-Net and ParametricNet by 21\% and 15\%, respectively, on the L-dataset and by 17\% and 27\% on the G-dataset. The results show that NormNet has stronger learning and generalization abilities than other state-of-the-art methods.

\textbf{Evaluation on MultiScale dataset}. PPR-Net, a state-of-the-art OPE method for stacked scenes, is selected as a baseline. The results are shown in table \ref{tabel:multiscale}, the performance of NormNet outperforms PPR-Net on all objects by 21\% in the mAP. The results show that NormNet performs better than PPR-Net when facing objects with significant scale changes in the stacked scenarios. As shown in Fig. \ref{fig:fig_compare}, the brick and thinboard are challenge to PPR-Net, but NormNet performs well.

\subsection{Ablation Study}
In this part, we explore the influence of depth missing generation in our Sim-to-Real transfer pipeline on the Sil\'eane dataset and probe the effect of scale normalization on the MultiScale dataset. 

\textbf{Effect of depth missing generation}. The result is shown in Table \ref{tabel:ablation_SRT}. NormNet (w/o DMG) is trained on synthetic data from IPADataset \cite{kleebergerlarge} only with domain randomization, which decreases APs of all objects. The result shows that the depth missing generation is helpful in bridging the Sim-to-Real gap.

\begin{table}[t]
\caption{ablation study on depth missing generation. w/o, without}
\resizebox{\linewidth}{!}{
\begin{tabular}{@{}cccccccc@{}}
\toprule
Method           & Brick         & C.stick      & Pepper        & Bunny         & Tless20       & Tless22                 \\ \midrule
NormNet(w/o DMG)  & 0.73          & 0.99         & 0.98          & 0.98          & 0.90          & 0.91                    \\
\textbf{NormNet} & \textbf{0.83} & \textbf{1.0} & \textbf{0.99} & \textbf{0.99} & \textbf{0.93} & \textbf{0.93}  \\ \bottomrule
\end{tabular}
}
\label{tabel:ablation_SRT}
\end{table}

\begin{table}[t]
\caption{ablation study on scale normalization module. SN, scale normalization; AT, affine transformation; w/o, without}
\resizebox{\linewidth}{!}{
\begin{tabular}{@{}cccccccc@{}}
\toprule
Method           & Brick         & Tless20       & Tless29       & Thinboard     & Bunny         & tless22       & Ringscrew     \\ \midrule
NormNet(w/o SN)          & 0.47          & 0.75          & 0.85          & 0.42          & 0.93          & 0.91          & 0.91          \\
NormNet(w/o AT)  & 0.86          & \textbf{0.98} & \textbf{0.97} & \textbf{0.97} & \textbf{0.99} & \textbf{0.98} & \textbf{0.97} \\
\textbf{NormNet} & \textbf{0.95} & \textbf{0.98} & \textbf{0.97} & \textbf{0.97} & \textbf{0.99} & \textbf{0.98} & \textbf{0.97} \\ \bottomrule
\end{tabular}
}
\vspace{-15pt}
\label{tabel:ablation_SN}
\end{table}

 \textbf{Effect of scale normalization}. The result is shown in Table \ref{tabel:ablation_SN}. NormNet(w/o SN) removes the scale normalization module and NormNet(w/o AT) means that the affine transformation is not applied to the single-category point clouds before they are fed into the shared pose estimator. According to the results of  NormNet(w/o SN) and NormNet(w/o AT), semantic segmentation simplifies the stacked scene, which increases the APs of all objects significantly. The results of NormNet(w/o AT) and NormNet show that the affine transformation scales all objects into optimal scale range, which improves the AP of small objects, i.e., Brick.

\subsection{Real-world Experiment}

To demonstrate the effectiveness of NormNet in real-world stacked scenes, we selected three objects in the MultiScale dataset, i.e., Brick, Tless20 and Tless29. To train the NormNet, we annotated 3500 synthetic multi-category stacked scenes, of which 3150 are used for training while 350 scenes are used for testing. We adopt the same evaluation metric as Sec. \ref{section:dataset_metrics}. Quantitative evaluation shows that the NormNet achieves AP values of 0.96, 0.96, and 1.0 for Brick, Tless20 and Tless29. 

The experiment setup is shown in Fig. 12(c), we printed Brick, Tless20, Tless29 and randomly threw them into a bin. The point clouds were captured by a fixed Mech-Eye Pro M Enhanced camera in our experiment. As shown in Fig. 12(a), the multi-category stacked scene was built. The point cloud captured by the camera was input into NormNet after background subtraction and farthest point sampling. As shown in Fig. 12(b), NormNet was able to robustly estimate all graspable objects in the stacked scene, including the small objects, i.e., Brick.

\begin{figure}[t]	
	\centering
	\subfigure[Stacked scenes]{	
		\begin{minipage}[b]{0.30\columnwidth}
			\includegraphics[width=1.0\columnwidth, height=3.4cm]{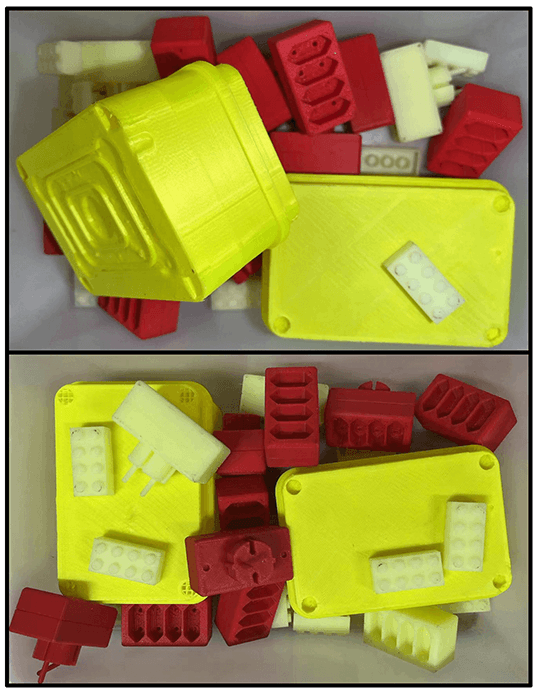}
		\end{minipage}
		\label{fig:fig9a}
		}
	\subfigure[Estimated poses]{\centering
		\begin{minipage}[b]{0.30\columnwidth}
			\includegraphics[width=1.0\columnwidth, height=3.4cm]{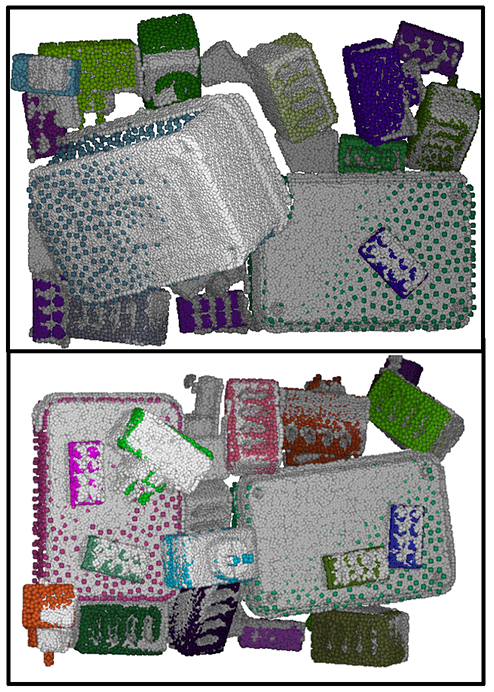}
		\end{minipage}
		\label{fig:fig9b}
		}
	\subfigure[Experiment setup]{\centering
		\begin{minipage}[b]{0.30\columnwidth}
			\includegraphics[width=1.0\columnwidth, height=3.4cm]{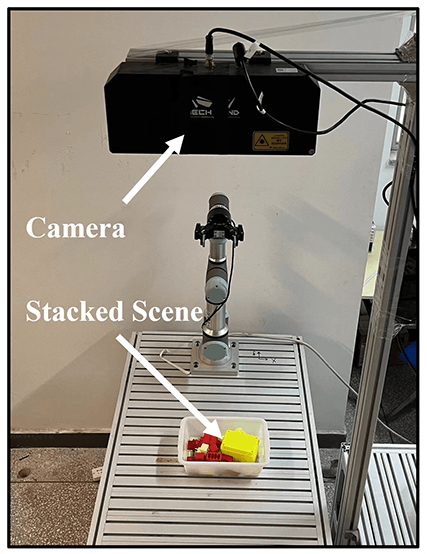}
		\end{minipage}
		\label{fig:fig9c}
		}					
	\caption{Experiment on real world. The estimated poses are overlapped with the point cloud.}
	\label{fig:fig9}
  \vspace{-10pt}
\end{figure}

\section{CONCLUSIONS}
To eliminate the impact of object scale, we proposed a new 6D pose estimation network NormNet which normalizes all objects into the same scale within the optimal scale range. To bridge the Sim-to-Real gap, we proposed a new Sim-to-Real pipeline based on style transfer and domain randomization. Compared with other state-of-the-art 6D OPE methods, Our method outperforms a large margin on benchmarks. We will release the code and dataset soon (\url{https://github.com/ShuttLeT/NormNet})

\clearpage
\bibliographystyle{IEEEtran}
\bibliography{ref}

\end{document}